\pdfoutput=1
\documentclass[runningheads]{llncs}

\usepackage{url}

\usepackage{amsmath}
\usepackage{amssymb}

\usepackage{cellspace}

\usepackage{pgfplots}

\usepackage{mathtools, nccmath}

\usepackage{tikz}
\usetikzlibrary{shapes}

\usepackage{algorithm}
\usepackage{algpseudocode}

\tikzset{
circ/.style={draw,circle,minimum height=3em},
}

\begin{document}
\setcounter{MaxMatrixCols}{50}

\title{PuzzlePoles: Cylindrical Fiducial Markers\\ Based on the PuzzleBoard Pattern}

\author{Juri Zach \and Peer Stelldinger\\
\authorrunning{J. Zach \and P. Stelldinger}
\institute{Faculty of Computer Science and Digital Society, HAW Hamburg, Germany} \email{\{juri.zach, peer.stelldinger\}@haw-hamburg.de}
}
\maketitle              % typeset the header of the contribution

\begin{abstract}
Reliable perception of the environment is a key enabler for autonomous systems, where calibration and localization tasks often rely on robust visual markers. We introduce the {\em PuzzlePole}, a new type of fiducial markers derived from the recently proposed PuzzleBoard calibration pattern. The PuzzlePole is a cylindrical marker, enabling reliable recognition and pose estimation from $360^\circ$ viewing direction. By leveraging the unique combinatorial structure of the PuzzleBoard pattern, PuzzlePoles provide a high accuracy in localization and orientation while being robust to occlusions. The design offers flexibility for deployment in diverse autonomous systems scenarios, ranging from robot navigation and SLAM to tangible interfaces.
\end{abstract}

%% \linenumbers

%% main text
\section{Introduction and Previous Work}

Accurate and reliable perception of the environment is crucial for autonomous systems to perform tasks such as navigation, mapping, localization, and object interaction. Fiducial markers have been widely adopted as artificial landmarks that provide well-defined geometric features for camera-based pose estimation and system calibration. These markers facilitate robust and efficient extraction of 2D-3D correspondences needed for localization algorithms and have been instrumental in advancing visual perception capabilities.

Among the most popular fiducial marker systems is \textbf{ArUco} \cite{garrido2014}, which consists of square binary patterns that can be quickly detected and decoded using a dictionary-based approach. ArUco markers are favored for their simplicity, low computational requirements, and ability to work reliably under various illumination and partial occlusion conditions. However, ArUco markers are planar and typically inflexible in terms of large viewpoint variations, often suffering degradation in pose estimation accuracy when viewed at oblique angles or from multiple directions \cite{kalaitzakis2021}. The same is true for similar planar fiducial markers like ARTag\cite{fiala2005} or AprilTag\cite{olson2011}.

For camera calibration purposes, \textbf{Charuco patterns} \cite{itseez2015} extend the ArUco concept by combining chessboard and ArUco markers, providing improved corner detection and enhanced geometric constraints leading to higher calibration accuracy. ArUco grids and Charuco markers are well-suited for camera calibration but maintain the limitation of planarity and directionality inherent to their flat design.

Beyond planar markers, 3D fiducial objects can potentially be detected from arbitrary directions. Beyond planar designs, 3D fiducial objects such as not only regular polyhedra \cite{garcia2023} but also cylindrical markers have been investigated: While some cylindrical markers are invariant regarding the rotation around the cylinder axis \cite{huelss2012}, others are restricted to a certain rotational sector \cite{jayarathne2013,zhang2017} or allow to detect the rotation for the whole 360° range \cite{wang2024}.
Wang et al.\cite{wang2024} place 12+ stripe-like "feature subregions" around the circumference of a cylinder in a de Bruijn-sequence-like encoding \cite{debruijn1946,herout2013} to define their CylinderTag. As at least two feature subregions need to be completely visible, their approach is not robust to partial occlusions. In contrast, Jayarathne et al.\cite{jayarathne2013} and Zhang et al.\cite{zhang2017} distribute local features on the cylinder surface to gain some robustness to partial occlusions. Both use chessboard corners as local features which Zhang et al. combine with circular dots. 
However, none of these approaches gives a 360° cylindrical encoding with occlusion robustness. A notable exception is an approach being presented just for comparison in the CylinderTag paper\cite{wang2024}: Wrapping a grid of standard 2D AprilTag fiducial markers around a cylinder obviously solves this task, but at the cost of the need of a high image resolution as a large number of individual tags needs to be decodable. Similarly, Xiao et al. present a 2d pattern which can be wrapped around a cylinder for optical tracking for tangible interaction and compare this to 2D ArUco markers on a cylinder surface \cite{xiao2025}. Again their solution, as well as the ArUco alternative, is robust to occlusions but requires some quite high resolution.

In order to be more robust regarding both occlusions and low resolution, one can distribute the local position encoding in overlapping areas by using so-called marker fields \cite{szentandrasi2012} or de Bruijn tori \cite{hurlbert1993,paterson1994,stelldinger2025}. Such an approach has been proposed by Zhu et al.\cite{zhu2023}, which has also been applied to cylindrical surfaces already \cite{cai2024}.

A similar and even more robust pattern is the recently proposed PuzzleBoard calibration pattern \cite{stelldinger2024}. The PuzzleBoard pattern presents a novel combinatorial design which integrates de Bruijn rings \cite{stelldinger2025} into a chessboard pattern. It permits high spatial encoding capacity and robust corner extraction. The PuzzleBoard pattern has been shown to outperform classical calibration patterns in terms of detection robustness and calibration accuracy both in case of severe occlusions and low image resolutions due to its ability to resolve ambiguities and provide rich spatial information.

\section{PuzzlePole: A Cylindrical Marker}

We leverage the PuzzleBoard pattern to introduce a new fiducial marker type:  \emph{PuzzlePole}, a cylindrical marker utilizing cyclic PuzzleBoard subpatterns to enable $360^\circ$ recognition and pose estimation. By using the combinatorial strength and spatial richness of the PuzzleBoard pattern we aim to address the shortcomings of existing cylindrical fiducial marker approaches. 

PuzzlePole markers are cylindrical poles wrapped with cyclic PuzzleBoard subpatterns. This configuration enables reliable detection and pose estimation from multiple viewing angles and directions, overcoming occlusions and viewpoint challenges typical for planar markers. 

The main advantage of the PuzzleBoard pattern compared to a grid of ArUco-like fiducial markers is that the position encoding of local patches is distributed on neighboring base cells, which allows decodability at much lower resolutions \cite{stelldinger2024}: Each patch of 3x3 puzzle pieces decodes a unique position by the configuration of its tabs and knobs. To identify tabs and knobs at low resolutions it is sufficient to check if the center between two corner points is black or white.

However, wrapping such a pattern around a cylinder in a seamless way requires periodicity of the code. Otherwise the code would be destroyed near the seam. Although the original PuzzleBoard pattern is periodic, it repeats only every 501 grid cells, which is nice for calibration purposes but disadvantageous when being used on a cylinder.

\begin{figure}[htbp]
    \centering
    \includegraphics[width=0.9\textwidth]{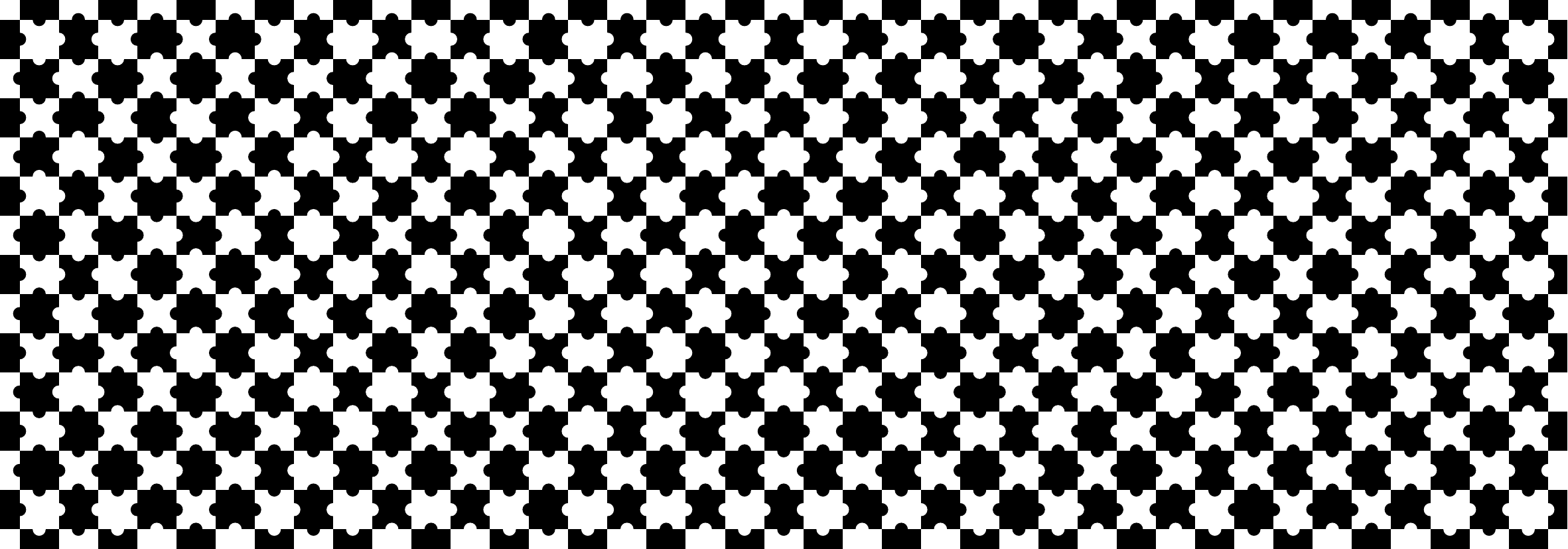}
    \caption{A sub-pattern of the PuzzleBoard pattern, which starts at $y=73$ (and $x=0$). As can be seen, the last row of puzzle pieces is identical to the first one. By wrapping the pattern onto a vertical cylinder and placing the first and last row exactly onto each other, one gets a seamless cyclic pattern.}
    \label{PuzzlePole:fig:puzzlepolecode}
\end{figure}

\begin{figure}[tp]
    \centering
    \includegraphics[width=0.99\textwidth]{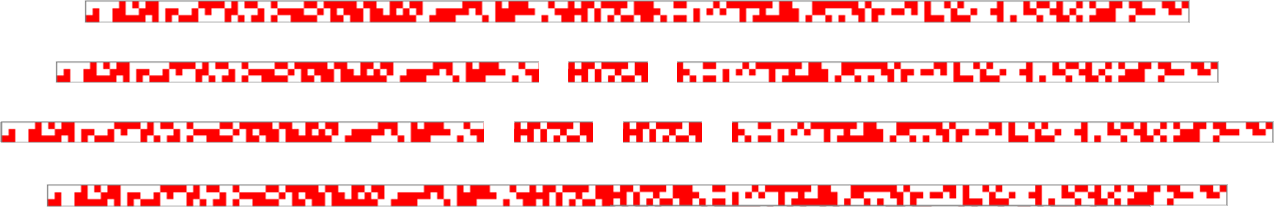}
    \caption{Repetition of a quasiperiodic sub-pattern. First row: The code stripe $B$ from the PuzzleBoard pattern\cite{stelldinger2024}. Second row: Separation of the stripe into three sub-patterns with the second (12 bit wide) sub-pattern being the quasi-periodic one starting at position 73. It starts with the same 2x6 bits as the following pattern. Third row: Repetition of the quasi-periodic sub-pattern. Fourth row: Stitching the sub-patterns together results in a pattern where every 3x3-sub-pattern does not generate new 3x3-sub-patterns at the stitching positions.}
    \label{PuzzlePole:fig:quasiperiodic}
\end{figure}

As a solution, one could generate new de Bruijn-like patterns with the appropriate period length, by using random uniform marker fields as in \cite{cai2024,herout2013,szentandrasi2012,zhu2023}. We choose another approach: We use quasiperiodic sub-patterns of the PuzzleBoard pattern, as this gives us a simple decoding algorithm with efficient error correction. Quasiperiodicity here means that the sub-pattern can be repeated twice within the whole pattern without generating new local 3-by-3 codes at the seams. As the PuzzleBoard pattern, as defined by Stelldinger et al.\cite{stelldinger2024}, is based on a periodic overlay of a horizontal code stripe of width 3 and height 167 (defining the bits on the vertical chessboard edges) with a vertical code stripe of width 167 and height 3 (defining the bits on the horizontal chessboard edges), one only needs to find 2 consecutive rows in the vertical stripe pattern, which exactly equal two other consecutive rows in the vertical stripe. If these two pairs of rows are a multiple of 6 apart from each other (stripe periodicity of 3 times chessboard periodicity of 2), then the corresponding two lines of puzzle pieces equal each other for the whole horizontal width of 501 puzzle pieces. Such a sub-pattern can be wrapped around a horizontal cylinder by overlapping the identical rows with each other. For example, generating a PuzzleBoard pattern using the Board Generator \cite{generator} setting \emph{start y} to $73$ and \emph{height} to $14$ (\emph{start x} and \emph{width} can be freely chosen), the first and the last row of generated puzzle pieces are equal, as shown in Figure \ref{PuzzlePole:fig:puzzlepolecode}. 
This corresponds to duplicating such a sub-pattern in the vertical code stripe of the PuzzleBoard, as illustrated in Figure \ref{PuzzlePole:fig:quasiperiodic}. In such a modified code stripe, only the internal local 3-by-3-patches of the sub-pattern are repeated twice, but no additional patch has been generated. By using such modified code stripes in the PuzzleBoard decoding algorithm, a PuzzlePole can seamlessly be decoded from any direction.

The same can be done for other row numbers than the shown 12 steps long section starting at $y=73$. Table \ref{PuzzlePole:tab:patterns} gives examples for different period lengths.

\begin{table}[h]
\centering
\begin{tabular}{|c|c|c|}
\hline
 \textbf{Period length} & \textbf{start y} & \textbf{height} \\
\hline
12 & 73 & 14  \\
\hline
18 & 7 & 20  \\
\hline
24 & 242 & 26  \\
\hline
30 & 176 & 32  \\
\hline
36 & 325 & 38  \\
\hline
42 & 410 & 44  \\
\hline
48 & 115 & 50  \\
\hline
\end{tabular}
\caption{Parameters for the PuzzleBoard Generator\cite{generator} to get cylindrical patterns with one piece overlap. These patterns can be wrapped around a cylinder to get a PuzzlePole marker. The period length is the number of puzzle pieces along the circumference of the cylinder.}
\label{PuzzlePole:tab:patterns}
\end{table}

\section{Detection Algorithm}\label{sec:detectionAlgorithm}

This section describes how a PuzzlePole can be detected within an image and localized in 3D-space relative to the camera.

We implemented a decoding algorithm for exactly the cylinder sizes given in Table  \ref{PuzzlePole:tab:patterns}. It allows to detect an arbitrary number of simultaneously visible PuzzlePoles. Decoding the pattern at the seam line where both ends of the cyclic sub-pattern join is realized by repeating the respective part of the corresponding code stripe.

The first steps of the detection algorithm are to (1) localize the chessboard corner points with subpixel accuracy in an image, (2) connect neighboring corner points to a grid, and (3) decode the grid in order to (4) identify each corner point, i.e. derive its unique id. This is being done by using our adaptation of the original PuzzleBoard detection algorithm \cite{generator}, as described in the previous section.

\subsection{Localization of a single PuzzlePole}

\begin{figure}[htp]
    \centering
    \includegraphics[width=0.8\textwidth]{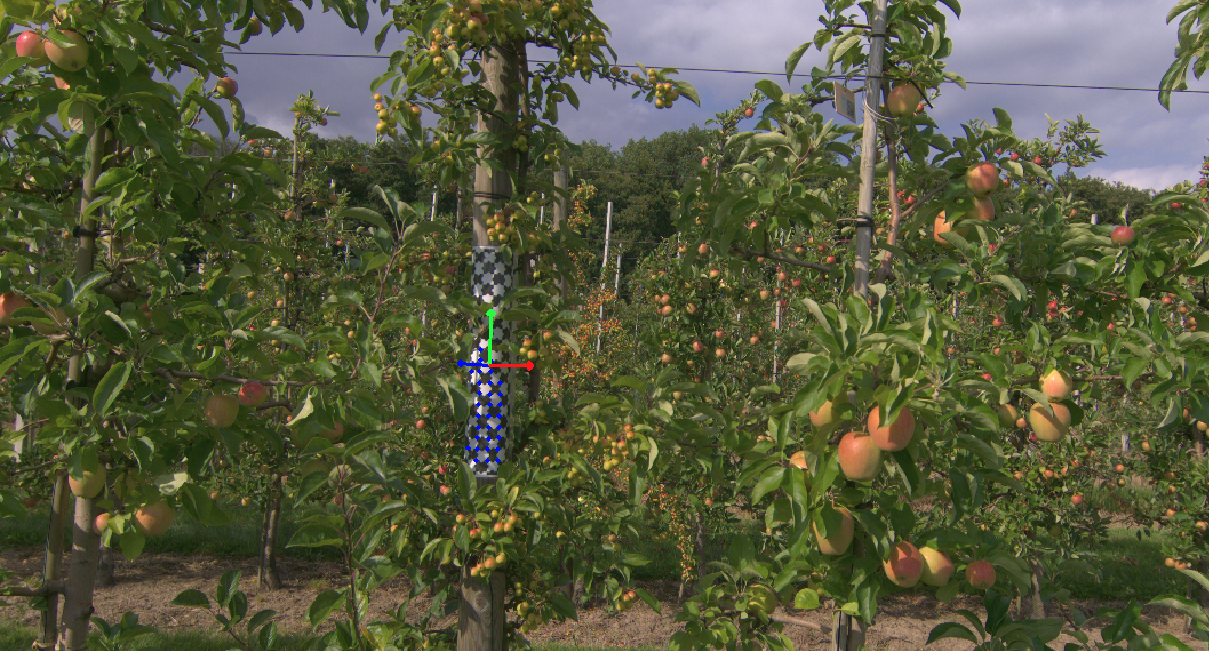}
    \caption{Localization of a partially occluded PuzzlePole}
    \label{PuzzlePole:fig:PuzzlePoleOccluded}
\end{figure}

To locate the exact position and rotation of a PuzzlePole, a calibrated camera and the section and size of the PuzzleBoard, from which the PuzzlePole is constructed, must be known.
In the first step, the PuzzleBoard detection algorithm detects the visible corner points of the PuzzleBoard Pattern.
Each corner point detected in the image has a unique ID.
Since the size and shape of the PuzzlePole is known, an ideal 3D-Model can be created, where each corner point has a 3D coordinate relative to the PuzzlePoles reference frame.
Due to the unique ID of each corner point, all visible points in the image can be mapped to the corresponding point in the 3D-Model of the PuzzlePole.
This results in a set of $n$ 2D and their corresponding 3D Points, which is a classical Perspective-$n$-Point (PnP) problem.
% TODO: how to solve a PnP Problem
To solve a PnP problem, the correct rotation and translation which maps a set 3D points onto the corresponding 2D point on an image plane is estimated.
The mathematical formulation is as follows:

\begin{equation}
    p_c = K [R|t] p_w
\end{equation}
Where $p_c=[u, v, 1]^T$ is the coordinate in the camera plane, $p_w = [x, y, z]^T$ is the 3D-world coordinate, $K$ is the intrinsic matrix of the camera and $R$ and $t$ are the rotation and translation matrices.
The PnP problem is well studied and several suitable solutions exist in public programming libraries like OpenCV. As a cylinder of known size has 6 degrees of freedom (3 for position and 3 for orientation), at least 3 well-distributed points on the cylinder surface need to be visible in the camera image for a unique solution, while a robust estimation should have at least 4 to 6 such points. As the detection of a PuzzleBoard pattern gives even more such points, the position and orientation of PuzzlePoles can be measured with high accuracy.
Figure \ref{PuzzlePole:fig:PuzzlePoleOccluded} shows the localization of a partially occluded PuzzlePole in an apple orchard.

\subsection{Localization of multiple PuzzlePoles}

\begin{figure}[htp]
    \centering
    \includegraphics[width=0.8\textwidth]{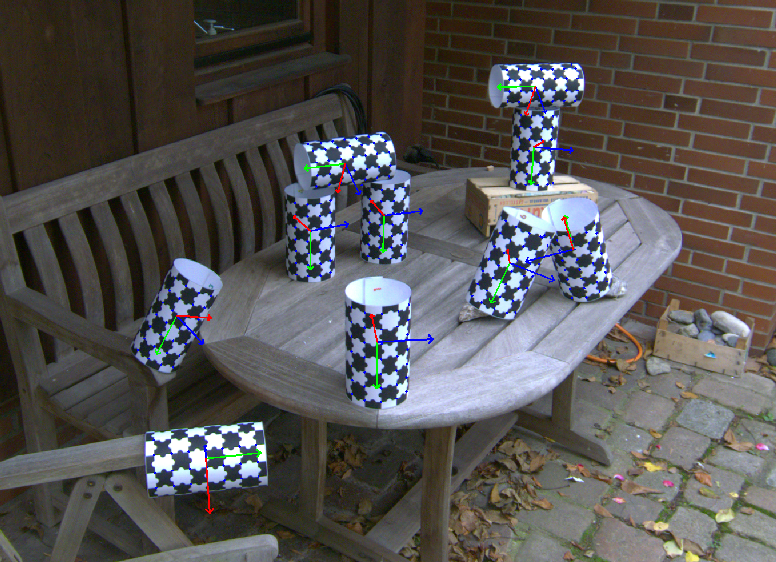}
    \caption{Localization of a multiple unique PuzzlePoles}
    \label{PuzzlePole:fig:manyPuzzlePoles}
\end{figure}

Depending on the application, multiple PuzzlePoles can be used at the same time. Figure \ref{PuzzlePole:fig:manyPuzzlePoles} shows the detection of multiple unique PuzzlePoles. Since the number of possible patterns from which a PuzzlePole can be created is very large (23 unique PuzzlePoles of the same size as given in Figure \ref{PuzzlePole:fig:PuzzlePoleOccluded} or 71 unique Poles of the size as given in Figure \ref{PuzzlePole:fig:manyPuzzlePoles}), the most simplest and robust solution is to create multiple unique PuzzlePoles, where each corner point can only belong to one PuzzlePole.
In this case, the corner points detected in an image can be assigned to the corresponding PuzzlePole and the PnP problem of each PuzzlePole is solved independently.

\section{Experiment: PuzzlePole localization accuracy}

The following experiment estimates the localization accuracy of the PuzzlePoles.
The experimental setup consists of two PuzzlePoles placed 2 meters apart with the same orientation.
A camera takes pictures of both PuzzlePoles from several angles and distances.
While it is difficult to measure the a ground truth for the camera's true distance and orientation to the PuzzlePoles for each position, the relative position and orientation between the two PuzzlePoles remain constant.
The position of both PuzzlePoles relative to the camera is estimated using the PuzzlePole localization algorithm described in Section \ref{sec:detectionAlgorithm}. These positions are then used to compute the relative position and orientation between the two PuzzlePoles.
By comparing the measured positions and orientations of the PuzzlePoles with the true values, the localization accuracy can be computed over many examples.\\

Images are taken with a Basler a2A5328-15ucPro industrial camera at a resolution of 2664 x 2304 Pixel and rectified using standard calibration methods.
The PuzzlePoles are constructed from a 7 x 12 periodic PuzzleBoard pattern with the $y$ corner point IDs from 73 to 85.
The edge length of a Puzzle piece is 3 centimeters. Thus, the PuzzlePoles each have a diameter of 11.46cm and a height (from lowest to highest corner point of the pattern) of 18cm.\\

\begin{table}[h]
\centering
\begin{tabular}{|c|c|c|c|}
\hline
 \textbf{Distance} & \textbf{Angle diff x} & \textbf{Angle diff y} & \textbf{Angle diff z} \\
\hline
2.002m & -1.65° & 1.15° & 1.10°  \\
\hline
\end{tabular}
\caption{Mean differences of the two PuzzlePoles in distance and orientation, measured by the localization algorithm.}
\label{PuzzlePole:tab:ExpMeanValues}
\end{table}

\begin{table}[h]
\centering
\begin{tabular}{|c|c|c|}
\hline
 \textbf{Measurement} & \textbf{Variance} & \textbf{Standard deviation} \\
\hline
Distance & 0.26 mm$^2$ & 16.26 mm  \\
Angle diff x & 7.20°$^2$ & 2.68° \\
Angle diff y & 1.68°$^2$ & 1.29° \\
Angle diff z & 7.26°$^2$ & 2.69° \\
\hline
\end{tabular}
\caption{Variance and standard deviation of distance and Rotation difference between the two PuzzlePoles.}
\label{PuzzlePole:tab:ExpVarStd}
\end{table}

\begin{figure}[htp]
    \centering
    \includegraphics[width=0.8\textwidth]{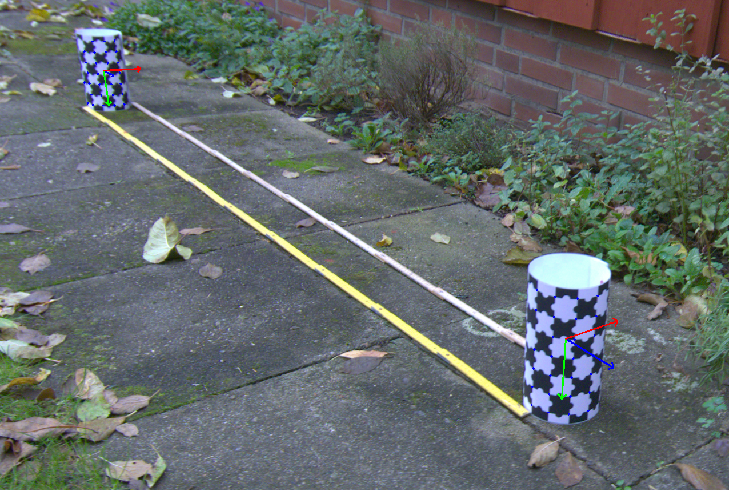}
    \caption{Experimental setup to measure the localization error between two PuzzlePoles, measured by the localization algorithm. The colors of the coordinate systems are: blue=x, green=y, red=z.}
    \label{PuzzlePole:fig:ExpSetup}
\end{figure}

\begin{figure}[htp]
    \centering
    \includegraphics[width=1.2\textwidth]{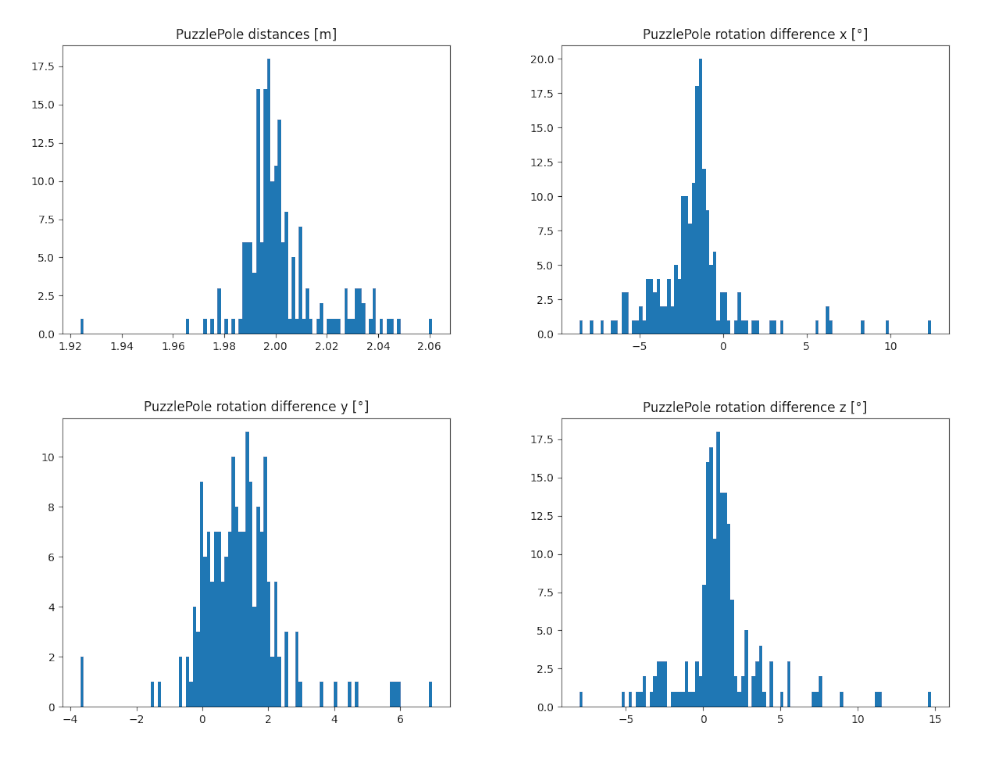}
    \caption{Histogram of the distance and orientation difference measurements between the two PuzzlePoles.}
    \label{PuzzlePole:fig:ExpHist}
\end{figure}

Measurements from 185 images along a trajectory spanning an angle of approximately 180° around the two PuzzlePoles have been acquired. Both poles have been visible in each of the images. The mean measurements results (see Table \ref{PuzzlePole:tab:ExpMeanValues}) show that the two PuzzlePoles have a distance of 2.002 meter (supposed distance is 2 meter) and an orientation difference of up to 1.65° (supposed orientation difference is zero).
These small differences between the expected and measured setups can be explained by the inaccuracy of the experimental setup. 
The results of the variance and standard deviation measurements (see Table \ref{PuzzlePole:tab:ExpVarStd}) show the noise of the localization algorithm.
The histogram of all measurements is shown in figure \ref{PuzzlePole:fig:ExpHist}.

The measurement results show that the position of a PuzzlePole can be located to an accuracy of a few centimeters. The rotation of the PuzzlePole can be estimated to within a few degrees. The accuracy of rotation around the cylinder center axis y (shown green in the figures) is twice that of the other two axes x (blue) and z (red). Since these are the standard deviations of the difference between two measurements (one for each pole), the measurements themselves deviate from the real orientation and position of the poles by a standard deviation which is even lower by a factor of $\sqrt{1/2}$.

When the distance between the detected corner points of the PuzzlePole in the image is compared with that of the ideal PuzzlePole model projected onto the image, a median distance of 0.37 pixels is reached.
This means that the localization algorithm is already achieving sub-pixel precision, indicating that a higher level of precision can mainly be achieved by increasing the image resolution of the PuzzlePole.

While the results of the above experiment give a good impression of the localization accuracy of PuzzlePoles, they are strongly dependent on the setup. A different camera, camera lens, PuzzlePole size or different lighting conditions can all lead to different results.

\section{Conclusion and Future Work}

This work presents the PuzzlePole, a new type of fiducial marker based on the PuzzleBoard pattern.
PuzzlePole characteristics include decoding at very low resolution and precise localizability for both relative position and rotation.
Depending on the use case, PuzzlePoles can be created in different sizes with a different number of puzzle pieces.
Due to the large number of possible PuzzleBoard sections that can be used to create a PuzzlePole, multiple PuzzlePoles can be used simultaneously and differentiated based on their corner point IDs.

In general, PuzzlePoles can be used to locate cylindrical objects relative to a camera, or vice versa.
Practical applications include localization and tracking of tools in manufacturing processes or medical devices, the use as a tangible user interface, robot localization and many more.
In future work, we will explore using PuzzlePoles as a loop closure point for robotic navigation.
Robots constantly need to track their position during operation.
While most robots measure their velocity and acceleration internally to keep track of their movement, these measurements are susceptible to drift and can result in large localization errors over time.
To solve this problem, external localization is required. Many robots rely on GPS signals or laser trackers; however, these systems are not available in all environments.
In the absence of any such systems, robots can use loop closure points. 
The robot recognizes features in the environment whose location is known and estimates its position relative to the loop closing point.

PuzzlePoles are ideal for loop closure due to their ease of decoding at low resolution and their high localization accuracy.
Follow-up studies will explore using PuzzlePoles for robot navigation in GPS-denied environments.
Initially, they will be employed as a validation method to assess the precision of internal robot localization techniques, such as visual odometry.
Subsequently, the internal robot localization techniques will be combined with loop closure using PuzzlePoles to investigate SLAM methods in such environments.

\section*{Acknowledgements}
This research was partially funded through project SAMSON (28DE201C21) by the Federal Ministry of Food and Agriculture (BMEL), based on a decision of the Parliament of the Federal Republic of Germany, via the Federal Office for Agriculture and Food (BLE) under the strategy for digitalization in agriculture.

\end{document}